\documentclass{article}

\usepackage{PRIMEarxiv}
\usepackage{lineno}
\usepackage{ulem}
\usepackage[style=authoryear,uniquename=false]{biblatex}

\usepackage{amsmath}
\usepackage{graphicx}
\usepackage{amssymb}
\usepackage{booktabs}
\usepackage{multirow}
\usepackage[colorlinks=true, allcolors=blue]{hyperref}
\usepackage{cleveref}

\crefname{figure}{Figure}{Figs.}
\crefname{table}{Table}{Tables}

\title{A review on vision-based motion estimation}
\author{
Hongyi Liu, Haifeng Wang\\
Washington State University \\
Pullman \\
\texttt{\{liu.hongyi, haifeng.wang\}@wsu.edu}
}

\begin{document}
\maketitle

\begin{abstract}
Compared to contact sensors-based motion measurement,
vision-based motion measurement has advantages of low cost and high efficiency
and have been under active development in the past decades. 
This paper provides a review on existing 
motion measurement methods. 
In addition to the development of each branch of vision-based motion 
measurement methods,
this paper also discussed the advantages and disadvantages of 
existing methods.
Based on this discussion, 
it was identified that existing methods have a common limitation 
in optimally balancing accuracy and robustness.
To address issue, 
we developed the Gaussian kernel-based motion measurement method.
Preliminary study shows that the developed method 
can achieve high accuracy on simple synthesized images. 
\end{abstract}


\section{Introduction}
Motion (vibration) of the structures is one of the most important 
signals to reflect the their states, which gains enduring attention in the 
past decades.
Contact and contactless sensors have been developed for motion measurement.
Contact sensors , such as accelerometer, linear 
potentiometer, and linear variable differential transformer (LVDT), 
can provide accurate measurement with high temporal resolution
and have been wide used. 
However, contact sensors face the  the challenge of high installation 
and maintenance cost
cost, especially for large-scale structures \parencite{khuc_completely_2017}.
Another drawback of
contact sensors is that they can only measure the motion at the installed
location,
thus generating a sparse motion field \parencite{yang_blind_2017}. 

With the advantage of efficient set up,
contactless motion measurement methods have been developed 
\parencite[e.g.,][]{paragios_optical_2006, li_motion_2014,li_motion_2022}. 
Among these contactless methods, 
vision-based measurement,
where the motion of the target object is measured from
a sequence of images,
has its unique advantage of 
low cost and dense measurement.
The vision-based motion measurement have been under active 
development in the past decades.

The vision-based motion measurement methods can be classified into two 
branches: matching-based methods and gradient-based methods 
\parencite{paragios_optical_2006, xu_review_2018}.
Matching-based methods extract motions by 
tracking the the motion of a template (e.g., a predefined image region)
in the image sequences
\parencite{weng_theory_1990, lowe_object_1999, leonardis_surf_2006},
Gradient-based methods extract motion information from 
pixel brightness variation
\parencite{lucas_iterative_1981,horn_determining_1981,fleet_computation_1990,
gautama_phase-based_2002}.
Matching-based methods are robust, 
however, it is typically difficult for matching-based methods to 
achieve high  (i.e., sub-pixel-level) accuracy since 
the motion of template is estimated in a pixel level.
On the other hand,
gradient-based methods can reach higher accuracy 
based on the subtle pixel brightness change caused by the 
motion of a target object.
While accurate,
gradient-based methods have 
are sensitive to the selection of
parameters.
As a result,
if a target object have different texture patterns, 
it is difficult for a gradient-based method with selected parameters
to reach high accuracy for all patterns,
i.e., low robustness. 

To simultaneously achieve high accuracy and high robustness,
we developed the Gaussian kernel-based motion measurement.
Specifically, 
each image is represented by a set of Gaussian kernels,
and the image sequence is represented by the position of 
these kernels.
This Gaussian kernel-based image representation is 
inspired by the 3D Gaussian splatting technique \parencite{kerbl_3d_2023}. 
3D Gaussian splatting
represents a 3D target objects by a set of 3D Gaussian kernels,
whose size, orientation, and color are optimized to 
fit the 3D view. 
Since 3D Gaussian splatting can 
can adaptively represent the object's features,
it has the potential to reach  high accuracy for all
patterns of an target object, i.e., high robustness.

The remainder of this paper is organized as follows. 
Section 2 provides a review for vision-based motion estimation.
To address the drawbacks of existing methods,
Section 3 introduces the Gaussian kernel-based motion estimation
with a preliminary study on synthetic images.
Section 4 draws the conclusions and points out the future work. 

\section{Review of vision-based motion estimation}

Existing vision-based motion estimation methods can be classified into two 
branches: matching-based methods, which work from Lagrangian perspective,
and 
gradient-based 
methods, which work from Eulerian perspective
\parencite{batchelor_introduction_2000,
liu_motion_2005,
wu_eulerian_2012,
wadhwa_phase-based_2013}.
In this section, the development of 
matching-based methods and gradient-based methods
are reviewed. 

\subsection{Matching-based methods}

Matching-based methods can be further categorized into two different
sub-categories: template-based matching and feature-based matching.
Template-based matching methods use a fixed block of pixels (template)
for matching. 
Features-based matching use image features, such as edges and corners,
for matching. 
Template matching is an intuitive motion 
method and will be introduced first.

\subsubsection{Template-based matching}

As for two sequential images, template matching (or block matching) method 
searches for 
the matching positions of a predefined "template"  in two sequential 
images, and 
use the position 
differences to estimate the target motion.
To search for the matching position,
it is necessary to quantify the similarity between the template and 
a searched area, i.e., a region in the image, with a metric.
\parencite{hashemi_template_2016}.
Various similarity metrics have been developed like 
mean square error (MSE), mean absolute difference (MAD), 
Peak to Signal Noise Ratio (PSNR) and sum of absolute difference (SAD) 
\parencite{khawase_overview_2017}.
By searching for the region that can reach a highest 
similarity metric, 
the  position of the template in the each image 
can be determined.

To find the position that can reach highest similarity metric,
a set of regions in the target image needs to be searched,
making the searching process time-consuming. 
To address this issue, 
efficient search methods have been developed 
For example,
three-step search (TSS)\parencite{koga_motion_1981},
four-step search (FSS)\parencite{po_novel_1996}, 
diamond search (DS)
, and 
hexagon-based search (HEXBS)
have been developed to increase the matching speed by optimizing the 
searching strategy.
\cite{singh_improved_2021} utilized particle swarm optimization to 
increase the matching speed in videos.

Template-matching methods inherently can only reach pixel-level accuracy.
Specifically, because of the template position is searched 
on pixel grids, 
the resolution of the estimated motions cannot reach sub-pixel-level. 
To reach a higher precision using template matching,
it is necessary to identify the matching position at sub-pixel level,
i.e., sub-pixel registration,
where the interpolation technique is one of the most used approach
\parencite{feng_vision-based_2015}.
The interpolation methods 
include bi-cubic
interpolation \parencite{choi_structural_2011}, second-order polynomial interpolation 
\parencite{sladek_development_2013}, edge-preserving interpolation 
\parencite{revaud_epicflow_2015}, etc. 
With these interpolation techniques, the accuracy of the estimated motions 
can reach up to 0.01–0.1 parts of the pixel \parencite{sladek_development_2013}.
In addition to interpolation, 
other approaches have also been developed,
including
phase correlation \parencite{reddy_fft-based_1996,ye_robust_2020}, 
Fourier-Mellin transformation\parencite{lei_novel_2009}
and deep learning \parencite{hikosaka_image--image_2022}.

Due to its simplicity and robustness, template 
matching is now commonly used as a part of 
hybrid methods to obtain robust, accurate and effective motion estimation 
methods. For example, \cite{gao_multiscale_2023} combined Kalman filter
with template matching to fuse
displacement and velocity estimation results to eliminate the influences of 
environment conditions and cumulative errors during displacement calculation
by integrating velocities. \cite{azimbeik_improved_2023} introduced 
camera calibration method into template matching, to improve the effectiveness
on large-scale structure and accuracy for low-amplitude motions. 
\cite{zheng_full-field_2024} used a lightweight slice model to improve the accuracy
of template matching on large slender ratio structures.

\subsubsection{Feature-based matching}

Despite numerous advantages, template matching has difficulties in 
considering
rotational motion, scale changes, illumination variation, 
occlusion, missing data, shade and background changes.
Feature-based matching has been developed to address these issues. 
Specifically, feature-based matchings use 
edges, corners,
and patches with interesting shapes \parencite{xu_review_2018, huang_survey_2024},
as features to track the motion.
Since these selected features are insensitive to changes in rotation, scale,
and illumination, 
feature-based matching can address the abovementioned difficulties.
In this subsection, the feature-based matching method is briefly reviewed. 

The core of feature-based matching is the feature descriptors, 
referring to the way to detect and describe features \parencite{leng_local_2019},
which have developed for decades from hand-crafted design to deep learning 
\parencite{chen_feature_2021}. 
For example, the
famous hand-crafted feature descriptor SIFT 
\parencite{lowe_object_1999} and its variant PCA-SIFT
\parencite{yan_ke_pca-sift_2004}, incorporating scale- and rotation- invariant 
property.
Improvements for on SIFT have been made
to achieve higher robustness and effiiency.
For example,
SURF speeds up the computation relying on integral images for image 
convolutions \parencite{leonardis_surf_2006}. 
DAISY attempts to avoid introducing
artifacts in dense flow calculation by wide-baseline image pairs
\parencite{tola_daisy_2010}. 
BRIEF uses binary string to further simplify the computation and thus 
improve the efficiency \parencite{hutchison_brief_2010}.
ORB works on BRIEF and adds the rotation-invariant property to it by adding corner 
orientation information into BRIEF descriptor\parencite{rublee_orb_2011}. 
BRISK improves the efficiency of SURF by adding a scale-space FAST-based 
detector \parencite{leutenegger_brisk_2011}.
KAZE processes features in a nonlinear scale space by 
nonlinear diffusion filtering to maintain more details of the raw 
images, and thus improve the matching accuracy\parencite{hutchison_kaze_2012}. 
FREAK uses a tighter and faster descriptor by binary strings from retinal 
sampling pattern, to save time and memory\parencite{alahi_freak_2012}.
With the development of deep learning,
more complex descriptors
based on deep learning have also been proposed: 
LIFT learns a pipeline for detection, orientation estimation, and feature 
description in a unified manner\parencite{yi_lift_2016}. 
L2Net learns a descriptor 
with good generalization ability based on Convolutional Neural Network 
(CNN)\parencite{tian_l2-net_2017}. 
HardNet improves the Lowe loss in SIFT, maximizing the distance 
between the closest positive and closest negative example in the 
batch\parencite{mishchuk_working_2017}. 
and GeoDesc integrates geometry constraints from multi-view 
reconstructions to obtain the superior ability on reconstruction
task \parencite{luo_geodesc_2018}. 

When it comes to special scenarios 
where the shape of the target is known,
some specifically designed methods can work based on the known shape to track 
and match. These predefined shapes include cross marker
\parencite{olaszek_investigation_1999},four-spot array \parencite{ho_synchronized_2012},
chessboard \parencite{ribeiro_non-contact_2014},
circle marker \parencite{chen_application_2015}, and
line-type structure \parencite{brownjohn_vision-based_2017}. 

For feature-based matching methods, 
the position description for features can reach a sub-pixel level.
However, 
ignores the pixel brightness variation information, 
and thus sacrifices the accuracy.
still concentrate on pixel-wise feature points, which means

\subsection{Gradient-based methods}

Gradient-based methods perform better than matching-based methods on flow 
density and accuracy by analysing the variation of intensity 
with the assumption that the intensity of a particular point in the pattern
is  constant before and after moving \parencite{horn_determining_1981, 
fleet_computation_1990}. 
There are two categories of gradient-based methods: 
Intensity-based optical flow and phase-based optical flow

\subsubsection{Intensity-based optical flow}

Lucas and Kanade (LK) optical flow

\parencite{lucas_iterative_1981}.
was proposed to 
find the motion information based on image brightness gradient.

With the assumptions that
the motion is small enough and the
brightness of a target point 
remains unchanged during motions
to maintain the

the constraint optical flow equation can be expressed as:
\begin{equation} 
    I_xu + I_yv + I_t = 0
\end{equation}
where $u$ and $v$ are the components of optical flow in $x$ and $y$ directions.
$I_x$,$I_y$ and $I_t$ are the respective partial derivatives of the image 
intensity, $I(x,y,t)$, with respect to horizontal direction $x$, vertical 
direction $y$ and time $t$.
Due to aperture problem, only one equation cannot 
determine the two unknown 
variables $u$ and $v$. LK optical flow 
assumed that the motions of nearby pixels
in a local region (e.g., a 3$\times$3 pixel region) are he same ,
and thus $u$ and $v$ can be optimally determined with 9 equations 
and just 2 unknowns.

Horn and Schunck (HS) proposed the similar method in a more widely 
used form in later variational optical flow methods with another additional 
constraint on the motions' gradients
\parencite{horn_determining_1981}. 
Compared to the LK optical flow that force nearby pixels 
to have the same motion,
the HS method makes the motion field of nearby pixels smooth.
Specifically, the HS method searches for a motion field (i.e., $u$ and $v$)
that can minimize the global energy function:

\begin{equation} 
    E(u,v) = E_D(u,v) + \lambda E_S(u,v) 
\end{equation}
with the data term:
\begin{equation} 
     E_D(u,v) = \int_\Omega (I(x+u,y+v,t+\Delta t)
    -I(x,y,t))^2 d\Omega 
    \label{eq:data_term}
\end{equation}
and a smoothness term as:
\begin{equation} 
     E_S(u,v) = 
    \int_\Omega \left[
    (\frac{\partial u}{\partial x})^2 + 
    (\frac{\partial u}{\partial y})^2 +
    (\frac{\partial v}{\partial x})^2 + 
    (\frac{\partial v}{\partial y})^2
    \right] d\Omega
    \label{eq:smoothness_ter}
\end{equation}
where $\lambda$ is the weight of smoothness term, $\Omega \subseteq \mathbb{R}^2$ 
represent the image domain.

Note that in HS optical flow, $u$ and $v$ 

are functions with variables $x$,$y$ and $t$.

Despite its advantages 

the HS optical flow is not capable to deal with motion discontinuities.
It's also difficult for HS optical flow to remain robust under
illumination changes and noise \parencite{tu_survey_2019}.
As a result, attempts have been made to  enhance the accuracy and improve the robustness
of HS optical flow in two directions: 
improved data term [Eq. (\ref{eq:data_term})]
and improved smoothness terms [Eq. (\ref{eq:smoothness_ter})].

For the data term,
\cite{kanade_high_2004,  zimmer_optic_2011}
added Gaussian filtering into image 
pre-processing stage to reduce the noise.
Other forms for data term were exploited to mitigate 
the influence of over-penalty on the local outliers 
\parencite{sun_secrets_2010, fortun_optical_2015,
monzon_regularization_2016, tu_survey_2019}, 
such as Charbonnier\parencite{bruhn_lucaskanade_2005,martorell_variational_2022, khan_nonlinear_2022}, 
Lorentzian\parencite{black_robust_1996, sun_secrets_2010}, 
Tukey\parencite{odobez_robust_1995}, Leclerc\parencite{memin_dense_1998}, 
and modified Hampel \parencite{senst_robust_2012}.
Illumination-robust consistency assumptions have also been introduced
in data term to
cope with illumination variations. For example, Brox et al. exploited an 
intensity gradient 
consistency term in data term to allow small intensity changes due to 
illumination \parencite{kanade_high_2004}.
\cite{papenberg_highly_2006} found the high-order
constraints the Hessian and the Laplacian could also improve the robustness
to illumination. Mohamed et al. proposed a 
texture-based consistency term for data term to handle large illumination
changes \parencite{mohamed_illumination-robust_2014}.

When it comes to the smoothness term, the improvements aimed for 
the ability to preserve edges and perform better on motion 
field with motion discontinuities. The first improvement was 
based on the same reason that the original $\mathcal{L}_2$ form of the smoothness
term in HS optical flow will overly penalize large motion gradient, and thus 
over-smooth the motion boundaries \parencite{black_robust_1996}. Some other 
smoothness term were also useful to preserve edges, 
such as the image-driven term in \parencite{alvarez_reliable_2000}, integrating 
intensity gradient into the smoothness term, the flow-driven term in 
\parencite{weickert_theoretical_2001}, the combined term in
\parencite{cremers_complementary_2009}.
Besides the ability to coping with motion discontinuities, 
modifications on smoothness term can also improve the accuracy of 
estimated motions by non-local regularization \parencite{sun_secrets_2010,
hutchison_efficient_2012},
extending the pixel's neighborhood,
and spatio-temporal 
regularization \parencite{zimmer_optic_2011, volz_modeling_2011}, considering temporal coherence 
on motion flows in temporal sequence. 

\subsubsection{Phased-based optical flow}

Another branch of gradient-based motion estimation approaches is phase-based 
optical flow, which converts intensity translation into phase shift based on 
Fourier shift theorem and connects the motion with phase shift.
\cite{fleet_computation_1990} pointed out these 
the phase contour is more robust to noise 
compared to image brightness.

In phase-based optical flow,
the phase information is extracted from the image sequence 
by a series of

directional

The phase contour motion 
could provide the accurate estimation of motions in the tuned direction of
the filter.
Consider the constant phase contour as:
\begin{equation} 
    \phi(x,y,t) = c, c \in R 
\end{equation}
By differentiating with respect to $t$, it transforms into the following form:
\begin{equation} 
    \phi_xu + \phi_yv + \phi_t = 0 
\end{equation}
where $\phi_x$,$\phi_y$ and $\phi_t$ are the respective partial derivatives of 
the filtered intensity signal's phase $\phi(x,y,t)$, with respect to horizontal 
direction $x$, vertical direction $y$ and time $t$, and $c$ is a constant. 
Since multiple filters with 
different directions can be applied for one pixel and its neighborhood, there's 
no need for phase-based method to add additional constraints to solve aperture 
problem.

\cite{gautama_phase-based_2002} improved phase-based method using spatial filters instead 
of spatial-temporal ones. They stated that,
instead of using the amplitude and frequency constraints to remove the possible 
unreliable measurement results, which was called instability, estimating the 
non-linearity of the phase-time pairs can give more stable results.
Meanwhile, 
the spatial filters can reduce the total number of filters by half and the 
computational cost due to the less dimension. However, if the practical 
motion is a vibration, the method to estimate the non-linearity of the temporal 
sequence of the phase signal is not appropriate because the temporal phase 
signal will not vary linearly.

\cite{chen_modal_2015} introduced phase-based method into structural vibration 
analysis. The filter bank, which aimed to cover
most frequency bands and
completely represented the image, was replaced with a single filter. The 
results of the 
frequency and the corresponding mode shape were extracted accurately. 

\cite{diamond_accuracy_2017} extended 
Chen's experiments to estimate the 
sub-pixel accuracy of phase-based methods in terms of vibration 
amplitude.
In this paper, they used one single filter to 
extract the displacements and estimated the potential accuracy of 
phase-based method, and picked the best results in the region of interest, 
leaving the active 
pixel selection, layer selection of image pyramid, selection of filter 
parameters in the practical application as an open question. Consequently, 
although in \parencite{diamond_accuracy_2017}, the accuracy of phase-based 
optical flow was ideally 
enough for many motion estimation applications, there was still a gap between this 
ideal accuracy 
in experiments and that of the practical applications.

To reduce the impact of manual parameter selection, researchers explored many 
directions. For example, \cite{liu_structural_2022} used a peak-picking technique to determine
the frequency range of a narrow Butterworth filter, instead of using a rough
scale of the moving target to determine the central frequency of a Gabor filter.
Then, for the approximate mono-component filtered signal, Hilbert transformation
was used to calculate the imaginary part of it to avoid additional parameter determination. 
\cite{miao_phase-based_2022}
explored the parameter selection including direction, standard deviation in two
directions, and the central frequency, on a simple stripe-like image sequence
to determine the optimal Gabor filter for this specific case. 
This parameter selection guidance was based on the stripe's direction, width and 
the "disturbing" gray pattern's width.

\subsubsection{Challenges of gradient-based motion estimation}

Despite numerous contributions on modifications and extensions on gradient-based
motion estimation methods, there are still many challenges like occlusions, 
over-smoothness,
large displacement, outliers, motion estimation on regions with no texture,
and rotation
\parencite{tu_survey_2019}. Also, the quality of gradient-based motion estimation
seriously depends on the parameter selection\parencite{ 
diamond_accuracy_2017,xu_review_2018}. For intensity-based motion estimation,
the specific term forms and the weight of each term can influence the results. For
phase-based motion estimation, the results are sensitive to the filter's parameters,
active pixel 
selection and the active pyramid level, and based on our current knowledge, there's 
no adaptive selection technique for general images. Both of the aforementioned cases
require the adaptive parameter determination.

\section{Gaussian splatting-base motion measurement}

Existing vision-based motion measurement methods suffer from the balance 
of accuracy and robustness. 
Specifically, 
matching-based methods have high robustness but low accuracy, 
especially on sub-pixel motions.
On the other hand, gradient-based methods can reach high accuracy, 
however, 
are sensitive to parameter determination and thus have low 
robustness.

In other words,
existing methods lack the ability to optimally balance accuracy and robustness. 
To address this issue,
we developed the adaptive Gaussian \textit{kernel-based image representation} 
technique based on
3D Gaussian splatting technique \parencite{kerbl_3d_2023}.
Using the position change of Gaussian kernels, 
the motion of an target object can be measured. 
The Gaussian kernel-based image representation is capable of 
reaching best fit for all regions in the target image,
and thus is capable of achieving the optimized balance between accuracy and robustness.

\subsection{2D Gaussian-splatting and corresponding motion estimation}

\subsubsection{3D Gaussian splatting}

With a set of initial 3D Gaussian 
kernels from Structure-from-Motion (SfM) and a fast render technique named 
elliptical weighted average (EWA) 
\parencite{schonberger_structure--motion_2016, zwicker_ewa_2002}, 3D Gaussian 
Splatting optimizes 
these kernels to reconstruct the whole 3D scene by minimizing the combined loss 
of $\mathcal{L}_1$ loss and SSIM between all the input 2D ground truth images 
$\{I_{gt}\}_M$ 
in different views and the corresponding rendered images $\{I_r\}_M$ 
\parencite{kerbl_3d_2023}.
Each pixel 
will obtain its rendered color by $\alpha$-blending method, which accumulates
the weighted colors from all the Gaussians' color $c_i$ in depth order along a 
ray as
\begin{equation} 
    C = \sum_{i=1}^{N} {T_i\alpha_ic_i} 
\end{equation}
with 
\begin{equation} 
    \alpha_i = o_i \text{exp} (-\frac{1}{2} (\mathbf{X}-\mathbf{\mu_i})^T 
    \mathbf{\Sigma}_i^{-1} (\mathbf{X}-\mathbf{\mu_i})) \; \text{and} \;
    T_i = \sum_{j=1}^{i-1} (1-\alpha_i) 
\end{equation}
where $o_i \in [0,1]$, $\mu_i \in \mathbb{R}^{3 \times 1}$, and 
$\Sigma_i \in \mathbb{R}^{3 \times 3}$ are the opacity, 3D mean and 3D 
covariance matrix of $i$-th Gaussian, respectively. Here, $\mathbf{X}$ 
represents the intersection coordinates between the ray and the $i$-th 3D
Gaussian kernel. 

\subsubsection{Development of 2D Gaussian kernel-based motion measurement }

The 3D Gaussian splatting-based scene reconstruction can 
achieve good visual results, 
however, 
cannot achieve high motion estimation accuracy from 
image sequences.
To achieve 2D Gaussian kernel-based motion estimation,
two aspects 
need to be considered: 
1) applying 3D Gaussian Splatting on 2D images,
and
2) using Gaussian kernels to represent motions between 
consecutive frames.

Applying 3D Gaussian Splatting to 2D images is necessary for 
Gaussian kernel-based motion estimation.
In this study, the 
2D Gaussian kernel-based image representation is developed 
based on the 3D Gaussian splatting technique:

\begin{equation} 
    C = \sum_{i=1}^{N} c_i^{'} \cdot \text{exp} (-\frac{1}{2} 
    (\mathbf{X}-\mathbf{\mu_i})^T 
    \mathbf{\Sigma}_i^{-1} (\mathbf{X}-\mathbf{\mu_i})) 
\end{equation}
where  $C$  is the color of a pixel,
$c_i^{'} \in \mathbb{R}^{3 \times 1}$ is the weighted color coefficient, 
$\mu _ i$ is the center coordinate vector for 2D Gaussian kernel $i$.

For Gaussian kernel-based motion measurement,
it is critical to link Gaussian kernels and motions.
As result, the Gaussian kernels should be consistent across
sequential images.

Based on the conventional optical flow's intensity consistency assumption 
\parencite{lucas_iterative_1981,horn_determining_1981}, local rigidity 
assumption from physically-based priors \parencite{ling_align_2024,
luiten_dynamic_2023}, and the two motion assumptions above, the local image 
patch will remain consistency in two consecutive frames.

In this case, the motion difference among kernels is used as the penalty term
to encourage

\subsection{Preliminary study on synthetic image.}

To verify the proposed Gaussian kernel-based motion estimation, we 
conducted a simple validation on synthetic image.

\subsubsection{Numerical validation setup}

\begin{figure}[h]
    \centering
    \includegraphics[scale = 0.7]{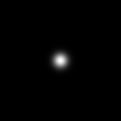} 
    \caption{The synthetic image for validation}
    \label{fig: synthesized image}
\end{figure}

\cref{fig: synthesized image} is the synthetic image for validation. It's a $121 \times 121$ 
grayscale image saved as 16-bit format. The central bright patch was generated 
by a Gaussian equation with known parameters in \cref{Table. 1} and then the 
intensity values was scaled to 16-bit color and rounded into integer intensity.

\begin{table}[h]
    \centering
    \caption{Gaussian kernel parameters of synthetic image for validation}
    \begin{tabular}{ll}
        \toprule
        Parameter & Value \\ \hline
        Horizontal standard deviation $\sigma_x$   & 4.8 pixels   \\
        Vertical standard deviation $\sigma_y$   & 4.8 pixels   \\ 
        Correlation coefficient $\rho$   & 0  \\ 
        Normalized intensity $c$   & 1.0  \\
        Normalized central position $(x,y)$  & (0,0)  \\
        \bottomrule
    \end{tabular}
    \label{Table. 1}
\end{table}

Because the image is synthesized based on selected parameters, we 
directly synthesize a second image with an  applied motion.
In this validation, we guided the central kernel to move 0.01 
pixels to the left and 0.01 pixels upward, and then repeated the process above 
to generate the second frame.

\subsubsection{Motion estimation process details}

The Gaussian kernel-based motion estimation process is composed of three steps. \\
Step 1: Initial Gaussian kernels were generated randomly (uniformly distributed 
in respective normalized range.). 
We remove
the initial Gaussian kernels at the locations that have unreasonably low brighness 
(i.e., brightness lower than $10^{-5}$). \\

Step 2, 
an $\mathcal{L}_1$-based parameter optimization process is performed to fit
the first frame (Fig. \ref{fig: synthesized image})
to a mean absolute error loss of $10^{-4}$. \\

Step 3, the parameter optimization process is performed to fit both frames with
loss function:
\begin{equation} 
    \mathcal{L} = \lambda_1 \mathcal{L}_1 + \lambda_2 \mathcal{L}_{\text{diff}} + %
    \lambda_3 \mathcal{L}_{\text{smooth}} 
\end{equation}
where $\lambda_1$, $\lambda_2$ and $\lambda_3$ are the corresponding weights for each 
loss term. In our test, they are 0.25, 0.25, and 0.50 respectively.

$\mathcal{L}_{\text{diff}}$ is added to constrain the representation quality in both frame:
\begin{equation}
    \mathcal{L}_{\text{diff}} = \| \mathcal{L}_1(I_1) - \mathcal{L}_1(I_2) \| 
\end{equation}
where $\mathcal{L}_1(I_1)$ and $\mathcal{L}_1(I_2)$ are the respective $\mathcal{L}_1$
losses for the first and second frames.
The motion smoothness loss term
$\mathcal{L}_{\text{smooth}}$ is also introduced to achieve a smooth motion field: 
\begin{equation}
    \mathcal{L}_{\text{smooth}} = \frac{\text{STD}(\mathbf{d}_i)}%
    {\| \text{MEAN}(\mathbf{d}_i)\|}
\end{equation}
where $\text{STD}(*)$ and $\text{MEAN}(*)$ are the standard deviation and mean value 
of the motions of all the kernels.

\subsubsection{Results and discussion}
\begin{figure}[h]
    \centering
    \includegraphics[scale=0.05]{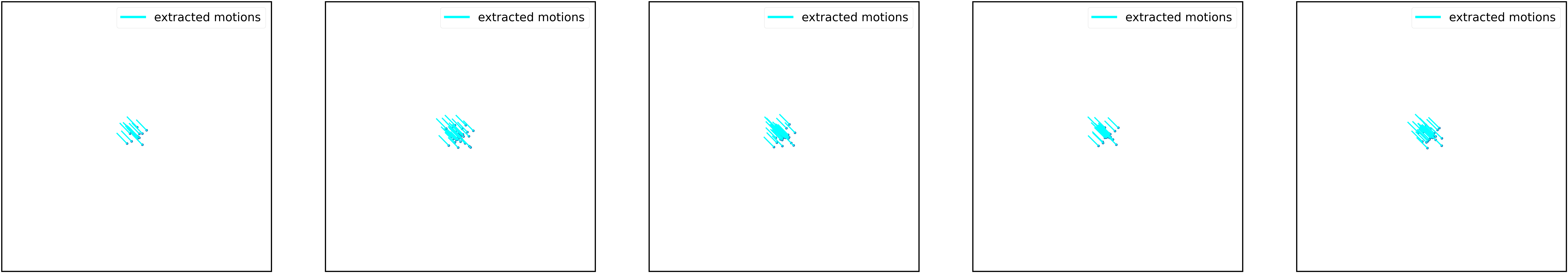} 
    \caption{Motion field for validation cases (Case1 - Case5, scaled by 500 times)}
    \label{Fig. 2}
\end{figure}

We used 5 different sets of initial kernels to show the robustness of the
Gaussian-Splatting-based motion estimation. \cref {Fig. 2} is the motion field and the
distribution of kernels for
each validation case and the distribution of the motion errors are shown in
\cref {Table. 2}.

\begin{table}[h]
    \centering
    \caption{Motion errors}
        \begin{tabular}{cccccccccc}
            \toprule
            \multirow{2}{*}{\textbf{Case}} 
            & \multirow{2}{*}{\shortstack{\textbf{Number of}\\ \textbf{Kernels}}}
            & \multicolumn{2}{c}{\shortstack{\textbf{Applied}\\ \textbf{Motions (pixel)}}}
            & \multicolumn{2}{c}{\shortstack{\textbf{Average}\\ \textbf{Errors (pixel)}}}
            & \multicolumn{2}{c}{\shortstack{\textbf{Relative}\\ \textbf{Average Errors (\%)}}}
            & \multicolumn{2}{c}{\shortstack{\textbf{Standard}\\ \textbf{Deviation ((pixel))}}}\\
            \cmidrule(lr){3-4}
            \cmidrule(lr){5-6}
            \cmidrule(lr){7-8}
            \cmidrule(lr){9-10}
            && \textbf{x} & \textbf{y} & \textbf{x} & \textbf{y} & \textbf{x} & \textbf{y}
            & \textbf{x} & \textbf{y}\\
            \midrule
            Case1  &30&0.01 & 0.01 &  3.10e-04 & 5.30e-05 &  3.10 & 0.53 &  1.12e-07 & 1.30e-07\\
            Case2  &58&0.01 & 0.01 &  6.21e-04 & 5.05e-04 &  6.21 & 5.05 &  1.09e-07 & 1.93e-07\\
            Case3  &63&0.01 & 0.01 &  5.52e-04 & 4.62e-04 &  5.52 & 4.62 &  1.18e-07 & 9.82e-08\\
            Case4  &52&0.01 & 0.01 &  2.56e-04 & 3.57e-04 &  2.56 & 3.57 &  7.29e-08 & 1.29e-07\\
            Case5  &58&0.01 & 0.01 &  1.54e-04 & 2.63e-05 &  1.54 & 0.26 &  9.48e-08 & 1.06e-07\\
            \midrule
            Average&-&  -    & -     &  3.79e-04 & 2.81e-04 &  3.79 & 2.81 &  -        & -       \\
            Max.   &-& -     & -     &  6.21e-04 & 5.05e-04 &  6.21 & 5.05 &  -        & -       \\
            Min.   &-& -     & -     &  1.54e-04 & 2.63e-05 &  1.54 & 0.26 &  -        & -       \\
            \bottomrule
        \end{tabular}
    \label{Table. 2}
\end{table}

In all 5 cases, the Gaussian-Splatting-based motion estimation can reach a good 
accuracy. As for the applied sub-pixel motions of 0.01 pixels, the maximum average 
relative error in the 5 cases is 6.21\% on x-axis and 5.05\% on y-axis with an 
average level of 3.79\% on x-axis and 2.81\% on y-axis. Based on these results, 
Gaussian-Splatting-based motion estimation shows a promising potential. We can 
expect a better results with further improvements on different aspects, such as
initialization, optimization process, parameter selection techniques, and loss 
function.

\section{Conclusions}

With a history of more than 70 years, vision-based motion measurement has
consistently received attention and have been 
continuously improved to be more effective and accurate.
This papers reviews the development of vision-base motion measurement 
with analysis of advantages and disadvantages of each method category.
Existing methods have the difficulty achieving optimized balance between
accuracy and robustness.
Specifically,
matching-based methods have high robustness but low accuracy.
Gradient-based methods accurate but less robust.
To address this balance issue, 
we developed the Gaussian-kernel-based motion measurement method.
The preliminary study shows that the developed method is 
capable of measuring 0.01 pixel motion with an average error of $3\times 10^{-4}$ pixels.

While the developed Gaussian kernel-based motion measurement has demonstrated
a potential to accurately estimate motions on a simple case, there are 
a few aspects that needs more improvement.

\begin{itemize}
    \item Optimization on more complex images. Our validation was conducted only 
    on simple case. As for complex images, optimization will be more likely to 
    fail, which indicates further work to improve the robustness of the 
    optimization.
    \item Repeatability. The existing kernel is determined based on randomly 
    initialized parameters. However, as a measurement tool, we expect the 
    kernel is optimally determined with repeatability. We expect to address this 
    issue training a convolutionaa neural network for parameter initialization.
\end{itemize}

\printbibliography

@article{liu_structural_2022,
	title = {Structural motion estimation via Hilbert transform enhanced phase-based video processing},
	volume = {166},
	issn = {08883270},
	url = {https://linkinghub.elsevier.com/retrieve/pii/S0888327021007676},
	doi = {10.1016/j.ymssp.2021.108418},
	pages = {108418},
	journaltitle = {Mechanical Systems and Signal Processing},
	shortjournal = {Mechanical Systems and Signal Processing},
	author = {Liu, G. and Li, M.Z. and Mao, Z. and Yang, Q.S.},
	urldate = {2024-01-07},
	date = {2022-03},
	langid = {english},
}

@article{horn_determining_1981,
	title = {Determining optical flow},
	volume = {17},
	issn = {00043702},
	url = {https://linkinghub.elsevier.com/retrieve/pii/0004370281900242},
	doi = {10.1016/0004-3702(81)90024-2},
	pages = {185--203},
	number = {1},
	journaltitle = {Artificial Intelligence},
	shortjournal = {Artificial Intelligence},
	author = {Horn, Berthold K.P. and Schunck, Brian G.},
	urldate = {2023-11-27},
	date = {1981-08},
	langid = {english},
}

@article{lucas_iterative_1981,
	title = {An Iterative Image Registration Technique with an Application to Stereo Vision},
	pages = {674--679},
	journaltitle = {{IJCAI}'81: 7th international joint conference on Artificial intelligence},
	author = {Lucas, Bruce D and Kanade, Takeo},
	date = {1981-08},
}

@inproceedings{weng_theory_1990,
	location = {Osaka, Japan},
	title = {A Theory of Image Matching},
	isbn = {978-0-8186-2057-7},
	url = {http://ieeexplore.ieee.org/document/139520/},
	doi = {10.1109/ICCV.1990.139520},
	eventtitle = {[1990] Third International Conference on Computer Vision},
	pages = {200--209},
	booktitle = {[1990] Proceedings Third International Conference on Computer Vision},
	publisher = {{IEEE} Comput. Soc. Press},
	author = {Weng, J.},
	urldate = {2023-11-19},
	date = {1990},
}

@article{xu_review_2018,
	title = {Review of machine-vision based methodologies for displacement measurement in civil structures},
	volume = {8},
	issn = {2190-5452, 2190-5479},
	url = {http://link.springer.com/10.1007/s13349-017-0261-4},
	doi = {10.1007/s13349-017-0261-4},
	pages = {91--110},
	number = {1},
	journaltitle = {Journal of Civil Structural Health Monitoring},
	shortjournal = {J Civil Struct Health Monit},
	author = {Xu, Yan and Brownjohn, James M. W.},
	urldate = {2023-10-11},
	date = {2018-01},
	langid = {english},
}

@article{revaud_epicflow_2015,
	title = {{EpicFlow}: Edge-Preserving Interpolation of Correspondences for Optical Flow},
	rights = {{arXiv}.org perpetual, non-exclusive license},
	url = {https://arxiv.org/abs/1501.02565},
	doi = {10.48550/ARXIV.1501.02565},
	shorttitle = {{EpicFlow}},
	author = {Revaud, Jerome and Weinzaepfel, Philippe and Harchaoui, Zaid and Schmid, Cordelia},
	urldate = {2023-10-02},
	date = {2015},
}

@article{miao_phase-based_2022,
	title = {Phase-based displacement measurement on a straight edge using an optimal complex Gabor filter},
	volume = {164},
	issn = {08883270},
	url = {https://linkinghub.elsevier.com/retrieve/pii/S0888327021005975},
	doi = {10.1016/j.ymssp.2021.108224},
	pages = {108224},
	journaltitle = {Mechanical Systems and Signal Processing},
	shortjournal = {Mechanical Systems and Signal Processing},
	author = {Miao, Yinan and Jeon, Jun Young and Kong, Yeseul and Park, Gyuhae},
	urldate = {2023-09-27},
	date = {2022-02},
	langid = {english},
}

@article{yang_blind_2017,
	title = {Blind Identification of Full-Field Vibration Modes from Video Measurements with Phase-Based Video Motion Magnification},
	volume = {85},
	issn = {08883270},
	url = {https://linkinghub.elsevier.com/retrieve/pii/S0888327016303272},
	doi = {10.1016/j.ymssp.2016.08.041},
	pages = {567--590},
	journaltitle = {Mechanical Systems and Signal Processing},
	shortjournal = {Mechanical Systems and Signal Processing},
	author = {Yang, Yongchao and Dorn, Charles and Mancini, Tyler and Talken, Zachary and Kenyon, Garrett and Farrar, Charles and Mascareñas, David},
	urldate = {2023-09-27},
	date = {2017-02},
	langid = {english},
}

@article{gautama_phase-based_2002,
	title = {A phase-based approach to the estimation of the optical flow field using spatial filtering},
	volume = {13},
	issn = {1045-9227},
	url = {http://ieeexplore.ieee.org/document/1031944/},
	doi = {10.1109/TNN.2002.1031944},
	pages = {1127--1136},
	number = {5},
	journaltitle = {{IEEE} Transactions on Neural Networks},
	shortjournal = {{IEEE} Trans. Neural Netw.},
	author = {Gautama, T. and Van Hulle, M.A.},
	urldate = {2023-09-27},
	date = {2002-09},
	langid = {english},
}

@article{fleet_computation_1990,
	title = {Computation of component image velocity from local phase information},
	volume = {5},
	issn = {0920-5691, 1573-1405},
	url = {http://link.springer.com/10.1007/BF00056772},
	doi = {10.1007/BF00056772},
	pages = {77--104},
	number = {1},
	journaltitle = {International Journal of Computer Vision},
	shortjournal = {Int J Comput Vision},
	author = {Fleet, David J. and Jepson, Allan D.},
	urldate = {2023-09-27},
	date = {1990-08},
	langid = {english},
}

@article{chen_modal_2015,
	title = {Modal identification of simple structures with high-speed video using motion magnification},
	volume = {345},
	issn = {0022460X},
	url = {https://linkinghub.elsevier.com/retrieve/pii/S0022460X1500070X},
	doi = {10.1016/j.jsv.2015.01.024},
	pages = {58--71},
	journaltitle = {Journal of Sound and Vibration},
	shortjournal = {Journal of Sound and Vibration},
	author = {Chen, Justin G. and Wadhwa, Neal and Cha, Young-Jin and Durand, Frédo and Freeman, William T. and Buyukozturk, Oral},
	urldate = {2023-09-27},
	date = {2015-06},
	langid = {english},
}

@article{wadhwa_phase-based_2013,
	title = {Phase-based video motion processing},
	volume = {32},
	issn = {0730-0301, 1557-7368},
	url = {https://dl.acm.org/doi/10.1145/2461912.2461966},
	doi = {10.1145/2461912.2461966},
	pages = {1--10},
	number = {4},
	journaltitle = {{ACM} Transactions on Graphics},
	shortjournal = {{ACM} Trans. Graph.},
	author = {Wadhwa, Neal and Rubinstein, Michael and Durand, Frédo and Freeman, William T.},
	urldate = {2023-09-26},
	date = {2013-07-21},
	langid = {english},
}

@article{diamond_accuracy_2017,
	title = {Accuracy Evaluation of Sub-Pixel Structural Vibration Measurements Through Optical Flow Analysis of a Video Sequence},
	volume = {95},
	issn = {02632241},
	url = {https://linkinghub.elsevier.com/retrieve/pii/S0263224116305693},
	doi = {10.1016/j.measurement.2016.10.021},
	pages = {166--172},
	journaltitle = {Measurement},
	shortjournal = {Measurement},
	author = {Diamond, D.H. and Heyns, P.S. and Oberholster, A.J.},
	urldate = {2023-05-22},
	date = {2017-01},
	langid = {english},
}

@article{kerbl_3d_2023,
	title = {3D Gaussian Splatting for Real-Time Radiance Field Rendering},
	volume = {42},
	issn = {0730-0301, 1557-7368},
	url = {https://dl.acm.org/doi/10.1145/3592433},
	doi = {10.1145/3592433},
	pages = {1--14},
	number = {4},
	journaltitle = {{ACM} Transactions on Graphics},
	shortjournal = {{ACM} Trans. Graph.},
	author = {Kerbl, Bernhard and Kopanas, Georgios and Leimkuehler, Thomas and Drettakis, George},
	urldate = {2023-12-21},
	date = {2023-08},
	langid = {english},
}

@inproceedings{liu_motion_2005,
	location = {Los Angeles California},
	title = {Motion magnification},
	isbn = {978-1-4503-7825-3},
	url = {https://dl.acm.org/doi/10.1145/1186822.1073223},
	doi = {10.1145/1186822.1073223},
	eventtitle = {{SIGGRAPH}05: Special Interest Group on Computer Graphics and Interactive Techniques Conference},
	pages = {519--526},
	booktitle = {{ACM} {SIGGRAPH} 2005 Papers},
	publisher = {{ACM}},
	author = {Liu, C. and Torralba, Antonio and Freeman, William T. and Durand, Frédo and Adelson, Edward H.},
	urldate = {2024-07-11},
	date = {2005-07},
	langid = {english},
}

@article{wu_eulerian_2012,
	title = {Eulerian video magnification for revealing subtle changes in the world},
	volume = {31},
	issn = {0730-0301, 1557-7368},
	url = {https://dl.acm.org/doi/10.1145/2185520.2185561},
	doi = {10.1145/2185520.2185561},
	pages = {1--8},
	number = {4},
	journaltitle = {{ACM} Transactions on Graphics},
	shortjournal = {{ACM} Trans. Graph.},
	author = {Wu, Hao-Yu and Rubinstein, Michael and Shih, Eugene and Guttag, John and Durand, Frédo and Freeman, William},
	urldate = {2024-07-11},
	date = {2012-08-05},
	langid = {english},
}

@incollection{leonardis_surf_2006,
	location = {Berlin, Heidelberg},
	title = {{SURF}: Speeded Up Robust Features},
	volume = {3951},
	rights = {http://www.springer.com/tdm},
	isbn = {978-3-540-33832-1 978-3-540-33833-8},
	url = {http://link.springer.com/10.1007/11744023_32},
	shorttitle = {{SURF}},
	pages = {404--417},
	booktitle = {Computer Vision – {ECCV} 2006},
	publisher = {Springer Berlin Heidelberg},
	author = {Bay, Herbert and Tuytelaars, Tinne and Van Gool, Luc},
	editor = {Leonardis, Aleš and Bischof, Horst and Pinz, Axel},
	urldate = {2024-07-12},
	date = {2006},
	langid = {english},
	doi = {10.1007/11744023_32},
	note = {Series Title: Lecture Notes in Computer Science},
}

@inproceedings{lowe_object_1999,
	location = {Kerkyra, Greece},
	title = {Object recognition from local scale-invariant features},
	isbn = {978-0-7695-0164-2},
	url = {http://ieeexplore.ieee.org/document/790410/},
	doi = {10.1109/ICCV.1999.790410},
	eventtitle = {Proceedings of the Seventh {IEEE} International Conference on Computer Vision},
	pages = {1150--1157 vol.2},
	booktitle = {Proceedings of the Seventh {IEEE} International Conference on Computer Vision},
	publisher = {{IEEE}},
	author = {Lowe, D.G.},
	urldate = {2024-07-12},
	date = {1999},
}

@misc{luiten_dynamic_2023,
	title = {Dynamic 3D Gaussians: Tracking by Persistent Dynamic View Synthesis},
	url = {http://arxiv.org/abs/2308.09713},
	shorttitle = {Dynamic 3D Gaussians},
	number = {{arXiv}:2308.09713},
	publisher = {{arXiv}},
	author = {Luiten, Jonathon and Kopanas, Georgios and Leibe, Bastian and Ramanan, Deva},
	urldate = {2024-07-12},
	date = {2023-08-18},
	eprinttype = {arxiv},
	eprint = {2308.09713 [cs]},
}

@inproceedings{sun_secrets_2010,
	location = {San Francisco, {CA}, {USA}},
	title = {Secrets of optical flow estimation and their principles},
	isbn = {978-1-4244-6984-0},
	url = {http://ieeexplore.ieee.org/document/5539939/},
	doi = {10.1109/CVPR.2010.5539939},
	eventtitle = {2010 {IEEE} Conference on Computer Vision and Pattern Recognition ({CVPR})},
	pages = {2432--2439},
	booktitle = {2010 {IEEE} Computer Society Conference on Computer Vision and Pattern Recognition},
	publisher = {{IEEE}},
	author = {Sun, Deqing and Roth, Stefan and Black, Michael J.},
	urldate = {2024-07-11},
	date = {2010-06},
}

@article{black_robust_1996,
	title = {The Robust Estimation of Multiple Motions: Parametric and Piecewise-Smooth Flow Fields},
	volume = {63},
	rights = {https://www.elsevier.com/tdm/userlicense/1.0/},
	issn = {10773142},
	url = {https://linkinghub.elsevier.com/retrieve/pii/S1077314296900065},
	doi = {10.1006/cviu.1996.0006},
	shorttitle = {The Robust Estimation of Multiple Motions},
	pages = {75--104},
	number = {1},
	journaltitle = {Computer Vision and Image Understanding},
	shortjournal = {Computer Vision and Image Understanding},
	author = {Black, Michael J. and Anandan, P.},
	urldate = {2024-07-11},
	date = {1996-01},
	langid = {english},
}

@article{li_motion_2014,
	title = {Motion Field Estimation for a Dynamic Scene Using a 3D {LiDAR}},
	volume = {14},
	rights = {https://creativecommons.org/licenses/by/3.0/},
	issn = {1424-8220},
	url = {https://www.mdpi.com/1424-8220/14/9/16672},
	doi = {10.3390/s140916672},
	pages = {16672--16691},
	number = {9},
	journaltitle = {Sensors},
	shortjournal = {Sensors},
	author = {Li, Qingquan and Zhang, Liang and Mao, Qingzhou and Zou, Qin and Zhang, Pin and Feng, Shaojun and Ochieng, Washington},
	urldate = {2024-07-16},
	date = {2014-09-09},
	langid = {english},
}

@article{li_motion_2022,
	title = {Motion Estimation and Coding Structure for Inter-Prediction of {LiDAR} Point Cloud Geometry},
	volume = {24},
	rights = {https://ieeexplore.ieee.org/Xplorehelp/downloads/license-information/{IEEE}.html},
	issn = {1520-9210, 1941-0077},
	url = {https://ieeexplore.ieee.org/document/9582802/},
	doi = {10.1109/TMM.2021.3119872},
	pages = {4504--4513},
	journaltitle = {{IEEE} Transactions on Multimedia},
	shortjournal = {{IEEE} Trans. Multimedia},
	author = {Li, Li and Li, Zhu and Liu, Shan and Li, Houqiang},
	urldate = {2024-07-16},
	date = {2022},
}

@incollection{paragios_optical_2006,
	location = {New York},
	title = {Optical Flow Estimation},
	isbn = {978-0-387-26371-7},
	url = {http://link.springer.com/10.1007/0-387-28831-7_15},
	pages = {237--257},
	booktitle = {Handbook of Mathematical Models in Computer Vision},
	publisher = {Springer-Verlag},
	author = {Fleet, D. and Weiss, Y.},
	editor = {Paragios, Nikos and Chen, Yunmei and Faugeras, Olivier},
	urldate = {2024-07-16},
	date = {2006},
	langid = {english},
	doi = {10.1007/0-387-28831-7_15},
}

@inproceedings{leutenegger_brisk_2011,
	location = {Barcelona, Spain},
	title = {{BRISK}: Binary Robust invariant scalable keypoints},
	isbn = {978-1-4577-1102-2 978-1-4577-1101-5 978-1-4577-1100-8},
	url = {http://ieeexplore.ieee.org/document/6126542/},
	doi = {10.1109/ICCV.2011.6126542},
	shorttitle = {{BRISK}},
	eventtitle = {2011 {IEEE} International Conference on Computer Vision ({ICCV})},
	pages = {2548--2555},
	booktitle = {2011 International Conference on Computer Vision},
	publisher = {{IEEE}},
	author = {Leutenegger, Stefan and Chli, Margarita and Siegwart, Roland Y.},
	urldate = {2024-07-16},
	date = {2011-11},
}

@article{tola_daisy_2010,
	title = {{DAISY}: An Efficient Dense Descriptor Applied to Wide-Baseline Stereo},
	volume = {32},
	rights = {https://ieeexplore.ieee.org/Xplorehelp/downloads/license-information/{IEEE}.html},
	issn = {0162-8828},
	url = {http://ieeexplore.ieee.org/document/4815264/},
	doi = {10.1109/TPAMI.2009.77},
	shorttitle = {{DAISY}},
	pages = {815--830},
	number = {5},
	journaltitle = {{IEEE} Transactions on Pattern Analysis and Machine Intelligence},
	shortjournal = {{IEEE} Trans. Pattern Anal. Mach. Intell.},
	author = {Tola, E. and Lepetit, V. and Fua, P.},
	urldate = {2024-07-16},
	date = {2010-05},
}

@incollection{hutchison_brief_2010,
	location = {Berlin, Heidelberg},
	title = {{BRIEF}: Binary Robust Independent Elementary Features},
	volume = {6314},
	rights = {http://www.springer.com/tdm},
	isbn = {978-3-642-15560-4 978-3-642-15561-1},
	url = {http://link.springer.com/10.1007/978-3-642-15561-1_56},
	shorttitle = {{BRIEF}},
	pages = {778--792},
	booktitle = {Computer Vision – {ECCV} 2010},
	publisher = {Springer Berlin Heidelberg},
	author = {Calonder, Michael and Lepetit, Vincent and Strecha, Christoph and Fua, Pascal},
	editor = {Hutchison, David and Kanade, Takeo and Kittler, Josef and Kleinberg, Jon M. and Mattern, Friedemann and Mitchell, John C. and Naor, Moni and Nierstrasz, Oscar and Pandu Rangan, C. and Steffen, Bernhard and Sudan, Madhu and Terzopoulos, Demetri and Tygar, Doug and Vardi, Moshe Y. and Weikum, Gerhard and Daniilidis, Kostas and Maragos, Petros and Paragios, Nikos},
	urldate = {2024-07-16},
	date = {2010},
	langid = {english},
	doi = {10.1007/978-3-642-15561-1_56},
	note = {Series Title: Lecture Notes in Computer Science},
}

@incollection{hutchison_kaze_2012,
	location = {Berlin, Heidelberg},
	title = {{KAZE} Features},
	volume = {7577},
	rights = {http://www.springer.com/tdm},
	isbn = {978-3-642-33782-6 978-3-642-33783-3},
	url = {http://link.springer.com/10.1007/978-3-642-33783-3_16},
	pages = {214--227},
	booktitle = {Computer Vision – {ECCV} 2012},
	publisher = {Springer Berlin Heidelberg},
	author = {Alcantarilla, Pablo Fernández and Bartoli, Adrien and Davison, Andrew J.},
	editor = {Fitzgibbon, Andrew and Lazebnik, Svetlana and Perona, Pietro and Sato, Yoichi and Schmid, Cordelia},
	editorb = {Hutchison, David and Kanade, Takeo and Kittler, Josef and Kleinberg, Jon M. and Mattern, Friedemann and Mitchell, John C. and Naor, Moni and Nierstrasz, Oscar and Pandu Rangan, C. and Steffen, Bernhard and Sudan, Madhu and Terzopoulos, Demetri and Tygar, Doug and Vardi, Moshe Y. and Weikum, Gerhard},
	editorbtype = {redactor},
	urldate = {2024-07-16},
	date = {2012},
	langid = {english},
	doi = {10.1007/978-3-642-33783-3_16},
	note = {Series Title: Lecture Notes in Computer Science},
}

@article{chen_feature_2021,
	title = {Feature detection and description for image matching: from hand-crafted design to deep learning},
	volume = {24},
	issn = {1009-5020, 1993-5153},
	url = {https://www.tandfonline.com/doi/full/10.1080/10095020.2020.1843376},
	doi = {10.1080/10095020.2020.1843376},
	shorttitle = {Feature detection and description for image matching},
	pages = {58--74},
	number = {1},
	journaltitle = {Geo-spatial Information Science},
	shortjournal = {Geo-spatial Information Science},
	author = {Chen, Lin and Rottensteiner, Franz and Heipke, Christian},
	urldate = {2024-07-16},
	date = {2021-01-02},
	langid = {english},
}

@inproceedings{singh_improved_2021,
	location = {Raigarh, India},
	title = {Improved Block Matching Motion Estimation Technique using Modified Particle Swarm Optimization in Video Coding},
	rights = {https://ieeexplore.ieee.org/Xplorehelp/downloads/license-information/{IEEE}.html},
	isbn = {978-1-66542-237-6},
	url = {https://ieeexplore.ieee.org/document/9619265/},
	doi = {10.1109/ETI4.051663.2021.9619265},
	eventtitle = {2021 Emerging Trends in Industry 4.0 ({ETI} 4.0)},
	pages = {1--6},
	booktitle = {2021 Emerging Trends in Industry 4.0 ({ETI} 4.0)},
	publisher = {{IEEE}},
	author = {Singh, Deepak},
	urldate = {2024-07-16},
	date = {2021-05-19},
}

@inproceedings{rublee_orb_2011,
	location = {Barcelona, Spain},
	title = {{ORB}: An efficient alternative to {SIFT} or {SURF}},
	isbn = {978-1-4577-1102-2 978-1-4577-1101-5 978-1-4577-1100-8},
	url = {http://ieeexplore.ieee.org/document/6126544/},
	doi = {10.1109/ICCV.2011.6126544},
	shorttitle = {{ORB}},
	eventtitle = {2011 {IEEE} International Conference on Computer Vision ({ICCV})},
	pages = {2564--2571},
	booktitle = {2011 International Conference on Computer Vision},
	publisher = {{IEEE}},
	author = {Rublee, Ethan and Rabaud, Vincent and Konolige, Kurt and Bradski, Gary},
	urldate = {2024-07-16},
	date = {2011-11},
}

@inproceedings{alahi_freak_2012,
	location = {Providence, {RI}},
	title = {{FREAK}: Fast Retina Keypoint},
	isbn = {978-1-4673-1228-8 978-1-4673-1226-4 978-1-4673-1227-1},
	url = {http://ieeexplore.ieee.org/document/6247715/},
	doi = {10.1109/CVPR.2012.6247715},
	shorttitle = {{FREAK}},
	eventtitle = {2012 {IEEE} Conference on Computer Vision and Pattern Recognition ({CVPR})},
	pages = {510--517},
	booktitle = {2012 {IEEE} Conference on Computer Vision and Pattern Recognition},
	publisher = {{IEEE}},
	author = {Alahi, A. and Ortiz, R. and Vandergheynst, P.},
	urldate = {2024-07-16},
	date = {2012-06},
}

@misc{yi_lift_2016,
	title = {{LIFT}: Learned Invariant Feature Transform},
	rights = {{arXiv}.org perpetual, non-exclusive license},
	url = {https://arxiv.org/abs/1603.09114},
	doi = {10.48550/ARXIV.1603.09114},
	shorttitle = {{LIFT}},
	publisher = {{arXiv}},
	author = {Yi, Kwang Moo and Trulls, Eduard and Lepetit, Vincent and Fua, Pascal},
	urldate = {2024-07-16},
	date = {2016},
	note = {Version Number: 2},
}

@inproceedings{tian_l2-net_2017,
	location = {Honolulu, {HI}},
	title = {L2-Net: Deep Learning of Discriminative Patch Descriptor in Euclidean Space},
	isbn = {978-1-5386-0457-1},
	url = {http://ieeexplore.ieee.org/document/8100132/},
	doi = {10.1109/CVPR.2017.649},
	shorttitle = {L2-Net},
	eventtitle = {2017 {IEEE} Conference on Computer Vision and Pattern Recognition ({CVPR})},
	pages = {6128--6136},
	booktitle = {2017 {IEEE} Conference on Computer Vision and Pattern Recognition ({CVPR})},
	publisher = {{IEEE}},
	author = {Tian, Yurun and Fan, Bin and Wu, Fuchao},
	urldate = {2024-07-16},
	date = {2017-07},
}

@misc{mishchuk_working_2017,
	title = {Working hard to know your neighbor's margins: Local descriptor learning loss},
	rights = {{arXiv}.org perpetual, non-exclusive license},
	url = {https://arxiv.org/abs/1705.10872},
	doi = {10.48550/ARXIV.1705.10872},
	shorttitle = {Working hard to know your neighbor's margins},
	publisher = {{arXiv}},
	author = {Mishchuk, Anastasiya and Mishkin, Dmytro and Radenovic, Filip and Matas, Jiri},
	urldate = {2024-07-16},
	date = {2017},
	note = {Version Number: 4},
}

@article{luo_geodesc_2018,
	title = {{GeoDesc}: Learning Local Descriptors by Integrating Geometry Constraints},
	rights = {Creative Commons Attribution Non Commercial Share Alike 4.0 International},
	url = {https://arxiv.org/abs/1807.06294},
	doi = {10.48550/ARXIV.1807.06294},
	shorttitle = {{GeoDesc}},
	author = {Luo, Zixin and Shen, Tianwei and Zhou, Lei and Zhu, Siyu and Zhang, Runze and Yao, Yao and Fang, Tian and Quan, Long},
	urldate = {2024-07-16},
	date = {2018},
	note = {Publisher: {arXiv}
Version Number: 2},
}

@article{senst_robust_2012,
	title = {Robust Local Optical Flow for Feature Tracking},
	volume = {22},
	issn = {1051-8215, 1558-2205},
	url = {http://ieeexplore.ieee.org/document/6209407/},
	doi = {10.1109/TCSVT.2012.2202070},
	pages = {1377--1387},
	number = {9},
	journaltitle = {{IEEE} Transactions on Circuits and Systems for Video Technology},
	shortjournal = {{IEEE} Trans. Circuits Syst. Video Technol.},
	author = {Senst, Tobias and Eiselein, Volker and Sikora, Thomas},
	urldate = {2024-07-17},
	date = {2012-09},
}

@article{memin_dense_1998,
	title = {Dense estimation and object-based segmentation of the optical flow with robust techniques},
	volume = {7},
	rights = {https://ieeexplore.ieee.org/Xplorehelp/downloads/license-information/{IEEE}.html},
	issn = {10577149},
	url = {http://ieeexplore.ieee.org/document/668027/},
	doi = {10.1109/83.668027},
	pages = {703--719},
	number = {5},
	journaltitle = {{IEEE} Transactions on Image Processing},
	shortjournal = {{IEEE} Trans. on Image Process.},
	author = {Memin, E. and Perez, P.},
	urldate = {2024-07-17},
	date = {1998-05},
}

@article{odobez_robust_1995,
	title = {Robust Multiresolution Estimation of Parametric Motion Models},
	volume = {6},
	rights = {https://www.elsevier.com/tdm/userlicense/1.0/},
	issn = {10473203},
	url = {https://linkinghub.elsevier.com/retrieve/pii/S1047320385710292},
	doi = {10.1006/jvci.1995.1029},
	pages = {348--365},
	number = {4},
	journaltitle = {Journal of Visual Communication and Image Representation},
	shortjournal = {Journal of Visual Communication and Image Representation},
	author = {Odobez, J.M. and Bouthemy, P.},
	urldate = {2024-07-17},
	date = {1995-12},
	langid = {english},
}

@inproceedings{martorell_variational_2022,
	location = {Online Streaming, --- Select a Country ---},
	title = {Variational Temporal Optical Flow for Multi-exposure Video:},
	isbn = {978-989-758-555-5},
	url = {https://www.scitepress.org/DigitalLibrary/Link.aspx?doi=10.5220/0010908300003124},
	doi = {10.5220/0010908300003124},
	shorttitle = {Variational Temporal Optical Flow for Multi-exposure Video},
	eventtitle = {17th International Conference on Computer Vision Theory and Applications},
	pages = {666--673},
	booktitle = {Proceedings of the 17th International Joint Conference on Computer Vision, Imaging and Computer Graphics Theory and Applications},
	publisher = {{SCITEPRESS} - Science and Technology Publications},
	author = {Martorell, Onofre and Buades, Antoni},
	urldate = {2024-07-17},
	date = {2022},
}

@article{khan_nonlinear_2022,
	title = {A nonlinear modeling of fractional order based variational model in optical flow estimation},
	volume = {261},
	issn = {00304026},
	url = {https://linkinghub.elsevier.com/retrieve/pii/S0030402622004958},
	doi = {10.1016/j.ijleo.2022.169136},
	pages = {169136},
	journaltitle = {Optik},
	shortjournal = {Optik},
	author = {Khan, Muzammil and Kumar, Pushpendra},
	urldate = {2024-07-17},
	date = {2022-07},
	langid = {english},
}

@article{bruhn_lucaskanade_2005,
	title = {Lucas/Kanade Meets Horn/Schunck: Combining Local and Global Optic Flow Methods},
	volume = {61},
	issn = {0920-5691},
	url = {http://link.springer.com/10.1023/B:VISI.0000045324.43199.43},
	doi = {10.1023/B:VISI.0000045324.43199.43},
	shorttitle = {Lucas/Kanade Meets Horn/Schunck},
	pages = {1--21},
	number = {3},
	journaltitle = {International Journal of Computer Vision},
	shortjournal = {International Journal of Computer Vision},
	author = {Bruhn, Andrés and Weickert, Joachim and Schnörr, Christoph},
	urldate = {2024-07-17},
	date = {2005-02},
	langid = {english},
}

@article{zimmer_optic_2011,
	title = {Optic Flow in Harmony},
	volume = {93},
	rights = {http://www.springer.com/tdm},
	issn = {0920-5691, 1573-1405},
	url = {http://link.springer.com/10.1007/s11263-011-0422-6},
	doi = {10.1007/s11263-011-0422-6},
	pages = {368--388},
	number = {3},
	journaltitle = {International Journal of Computer Vision},
	shortjournal = {Int J Comput Vis},
	author = {Zimmer, Henning and Bruhn, Andrés and Weickert, Joachim},
	urldate = {2024-07-17},
	date = {2011-07},
	langid = {english},
}

@incollection{kanade_high_2004,
	location = {Berlin, Heidelberg},
	title = {High Accuracy Optical Flow Estimation Based on a Theory for Warping},
	volume = {3024},
	rights = {http://www.springer.com/tdm},
	isbn = {978-3-540-21981-1 978-3-540-24673-2},
	url = {http://link.springer.com/10.1007/978-3-540-24673-2_3},
	pages = {25--36},
	booktitle = {Computer Vision - {ECCV} 2004},
	publisher = {Springer Berlin Heidelberg},
	author = {Brox, Thomas and Bruhn, Andrés and Papenberg, Nils and Weickert, Joachim},
	editor = {Pajdla, Tomás and Matas, Jiří},
	editorb = {Kanade, Takeo and Kittler, Josef and Kleinberg, Jon M. and Mattern, Friedemann and Mitchell, John C. and Nierstrasz, Oscar and Pandu Rangan, C. and Steffen, Bernhard and Sudan, Madhu and Terzopoulos, Demetri and Tygar, Dough and Vardi, Moshe Y. and Weikum, Gerhard},
	editorbtype = {redactor},
	urldate = {2024-07-17},
	date = {2004},
	langid = {english},
	doi = {10.1007/978-3-540-24673-2_3},
	note = {Series Title: Lecture Notes in Computer Science},
}

@article{tu_survey_2019,
	title = {A survey of variational and {CNN}-based optical flow techniques},
	volume = {72},
	issn = {09235965},
	url = {https://linkinghub.elsevier.com/retrieve/pii/S0923596518302479},
	doi = {10.1016/j.image.2018.12.002},
	pages = {9--24},
	journaltitle = {Signal Processing: Image Communication},
	shortjournal = {Signal Processing: Image Communication},
	author = {Tu, Zhigang and Xie, Wei and Zhang, Dejun and Poppe, Ronald and Veltkamp, Remco C. and Li, Baoxin and Yuan, Junsong},
	urldate = {2024-07-17},
	date = {2019-03},
	langid = {english},
}

@article{ribeiro_non-contact_2014,
	title = {Non-contact measurement of the dynamic displacement of railway bridges using an advanced video-based system},
	volume = {75},
	issn = {01410296},
	url = {https://linkinghub.elsevier.com/retrieve/pii/S0141029614002958},
	doi = {10.1016/j.engstruct.2014.04.051},
	pages = {164--180},
	journaltitle = {Engineering Structures},
	shortjournal = {Engineering Structures},
	author = {Ribeiro, D. and Calçada, R. and Ferreira, J. and Martins, T.},
	urldate = {2024-07-17},
	date = {2014-09},
	langid = {english},
}

@article{ho_synchronized_2012,
	title = {A Synchronized Multipoint Vision-Based System for Displacement Measurement of Civil Infrastructures},
	volume = {2012},
	rights = {http://creativecommons.org/licenses/by/3.0/},
	issn = {1537-744X},
	url = {http://www.hindawi.com/journals/tswj/2012/519146/},
	doi = {10.1100/2012/519146},
	pages = {1--9},
	journaltitle = {The Scientific World Journal},
	shortjournal = {The Scientific World Journal},
	author = {Ho, Hoai-Nam and Lee, Jong-Han and Park, Young-Soo and Lee, Jong-Jae},
	urldate = {2024-07-17},
	date = {2012},
	langid = {english},
}

@article{olaszek_investigation_1999,
	title = {Investigation of the dynamic characteristic of bridge structures using a computer vision method},
	volume = {25},
	rights = {https://www.elsevier.com/tdm/userlicense/1.0/},
	issn = {02632241},
	url = {https://linkinghub.elsevier.com/retrieve/pii/S0263224199000068},
	doi = {10.1016/S0263-2241(99)00006-8},
	pages = {227--236},
	number = {3},
	journaltitle = {Measurement},
	shortjournal = {Measurement},
	author = {Olaszek, Piotr},
	urldate = {2024-07-17},
	date = {1999-04},
	langid = {english},
}

@article{chen_application_2015,
	title = {Application of digital photogrammetry techniques in identifying the mode shape ratios of stay cables with multiple camcorders},
	volume = {75},
	issn = {02632241},
	url = {https://linkinghub.elsevier.com/retrieve/pii/S0263224115003838},
	doi = {10.1016/j.measurement.2015.07.037},
	pages = {134--146},
	journaltitle = {Measurement},
	shortjournal = {Measurement},
	author = {Chen, Chien-Chou and Wu, Wen-Hwa and Tseng, Hong-Zeng and Chen, Chi-Hong and Lai, Gwolong},
	urldate = {2024-07-17},
	date = {2015-11},
	langid = {english},
}

@article{brownjohn_vision-based_2017,
	title = {Vision-Based Bridge Deformation Monitoring},
	volume = {3},
	issn = {2297-3362},
	url = {http://journal.frontiersin.org/article/10.3389/fbuil.2017.00023/full},
	doi = {10.3389/fbuil.2017.00023},
	journaltitle = {Frontiers in Built Environment},
	shortjournal = {Front. Built Environ.},
	author = {Brownjohn, James Mark William and Xu, Yan and Hester, David},
	urldate = {2024-07-17},
	date = {2017-04-19},
}

@article{sladek_development_2013,
	title = {Development of a Vision Based Deflection Measurement System and Its Accuracy Assessment},
	volume = {46},
	issn = {02632241},
	url = {https://linkinghub.elsevier.com/retrieve/pii/S0263224112003934},
	doi = {10.1016/j.measurement.2012.10.021},
	pages = {1237--1249},
	number = {3},
	journaltitle = {Measurement},
	shortjournal = {Measurement},
	author = {Sładek, Jerzy and Ostrowska, Ksenia and Kohut, Piotr and Holak, Krzysztof and Gąska, Adam and Uhl, Tadeusz},
	urldate = {2024-07-17},
	date = {2013-04},
	langid = {english},
}

@article{choi_structural_2011,
	title = {Structural dynamic displacement vision system using digital image processing},
	volume = {44},
	rights = {https://www.elsevier.com/tdm/userlicense/1.0/},
	issn = {09638695},
	url = {https://linkinghub.elsevier.com/retrieve/pii/S0963869511000764},
	doi = {10.1016/j.ndteint.2011.06.003},
	pages = {597--608},
	number = {7},
	journaltitle = {{NDT} \& E International},
	shortjournal = {{NDT} \& E International},
	author = {Choi, Hyoung-Suk and Cheung, Jin-Hwan and Kim, Sang-Hyo and Ahn, Jin-Hee},
	urldate = {2024-07-17},
	date = {2011-11},
	langid = {english},
}

@book{batchelor_introduction_2000,
	edition = {1},
	title = {An Introduction to Fluid Dynamics},
	rights = {https://www.cambridge.org/core/terms},
	isbn = {978-0-521-66396-0 978-0-511-80095-5},
	url = {https://www.cambridge.org/core/product/identifier/9780511800955/type/book},
	publisher = {Cambridge University Press},
	author = {Batchelor, G. K.},
	urldate = {2024-07-16},
	date = {2000-02-28},
	doi = {10.1017/CBO9780511800955},
}

@article{po_novel_1996,
	title = {A novel four-step search algorithm for fast block motion estimation},
	volume = {6},
	rights = {https://ieeexplore.ieee.org/Xplorehelp/downloads/license-information/{IEEE}.html},
	issn = {10518215},
	url = {http://ieeexplore.ieee.org/document/499840/},
	doi = {10.1109/76.499840},
	pages = {313--317},
	number = {3},
	journaltitle = {{IEEE} Transactions on Circuits and Systems for Video Technology},
	shortjournal = {{IEEE} Trans. Circuits Syst. Video Technol.},
	author = {Po, Lai-Man and Ma, Wing-Chung},
	urldate = {2024-07-16},
	date = {1996-06},
}

@inproceedings{khawase_overview_2017,
	title = {An Overview of Block Matching Algorithms for Motion Vector Estimation},
	url = {https://fedcsis.org/proceedings/rice2017/drp/85.html},
	doi = {10.15439/2017R85},
	eventtitle = {The Second International Conference on Research in Intelligent and Computing in Engineering},
	pages = {217--222},
	author = {Khawase, Sonam T. and Kamble, Shailesh D. and Thakur, Nileshsingh V. and Patharkar, Akshay S.},
	urldate = {2024-07-16},
	date = {2017-06-09},
}

@article{feng_vision-based_2015,
	title = {A Vision-Based Sensor for Noncontact Structural Displacement Measurement},
	volume = {15},
	rights = {https://creativecommons.org/licenses/by/4.0/},
	issn = {1424-8220},
	url = {https://www.mdpi.com/1424-8220/15/7/16557},
	doi = {10.3390/s150716557},
	pages = {16557--16575},
	number = {7},
	journaltitle = {Sensors},
	shortjournal = {Sensors},
	author = {Feng, Dongming and Feng, Maria and Ozer, Ekin and Fukuda, Yoshio},
	urldate = {2024-07-16},
	date = {2015-07-09},
	langid = {english},
}

@misc{hashemi_template_2016,
	title = {Template Matching Advances and Applications in Image Analysis},
	rights = {{arXiv}.org perpetual, non-exclusive license},
	url = {https://arxiv.org/abs/1610.07231},
	doi = {10.48550/ARXIV.1610.07231},
	publisher = {{arXiv}},
	author = {Hashemi, Nazanin Sadat and Aghdam, Roya Babaie and Ghiasi, Atieh Sadat Bayat and Fatemi, Parastoo},
	urldate = {2024-07-16},
	date = {2016},
	note = {Version Number: 1},
}

@article{papenberg_highly_2006,
	title = {Highly Accurate Optic Flow Computation with Theoretically Justified Warping},
	volume = {67},
	rights = {http://www.springer.com/tdm},
	issn = {0920-5691, 1573-1405},
	url = {http://link.springer.com/10.1007/s11263-005-3960-y},
	doi = {10.1007/s11263-005-3960-y},
	pages = {141--158},
	number = {2},
	journaltitle = {International Journal of Computer Vision},
	shortjournal = {Int J Comput Vision},
	author = {Papenberg, Nils and Bruhn, Andrés and Brox, Thomas and Didas, Stephan and Weickert, Joachim},
	urldate = {2024-07-17},
	date = {2006-04},
	langid = {english},
}

@article{mohamed_illumination-robust_2014,
	title = {Illumination-Robust Optical Flow Using a Local Directional Pattern},
	volume = {24},
	rights = {https://ieeexplore.ieee.org/Xplorehelp/downloads/license-information/{IEEE}.html},
	issn = {1051-8215, 1558-2205},
	url = {http://ieeexplore.ieee.org/document/6748891/},
	doi = {10.1109/TCSVT.2014.2308628},
	pages = {1499--1508},
	number = {9},
	journaltitle = {{IEEE} Transactions on Circuits and Systems for Video Technology},
	shortjournal = {{IEEE} Trans. Circuits Syst. Video Technol.},
	author = {Mohamed, Mahmoud A. and Rashwan, Hatem A. and Mertsching, Barbel and Garcia, Miguel Angel and Puig, Domenec},
	urldate = {2024-07-17},
	date = {2014-09},
}

@article{alvarez_reliable_2000,
	title = {Reliable Estimation of Dense Optical Flow Fields with Large Displacements},
	volume = {39},
	issn = {09205691},
	url = {http://link.springer.com/10.1023/A:1008170101536},
	doi = {10.1023/A:1008170101536},
	pages = {41--56},
	number = {1},
	journaltitle = {International Journal of Computer Vision},
	author = {Alvarez, Luis and Weickert, Joachim and Sánchez, Javier},
	urldate = {2024-07-17},
	date = {2000},
}

@article{weickert_theoretical_2001,
	title = {A Theoretical Framework for Convex Regularizers in {PDE}-Based Computation of Image Motion},
	volume = {45},
	issn = {09205691},
	url = {http://link.springer.com/10.1023/A:1013614317973},
	doi = {10.1023/A:1013614317973},
	pages = {245--264},
	number = {3},
	journaltitle = {International Journal of Computer Vision},
	author = {Weickert, Joachim and Schnörr, Christoph},
	urldate = {2024-07-17},
	date = {2001},
}

\end{document}